\documentclass[letterpaper, 10 pt, conference]{ieeeconf}  

\IEEEoverridecommandlockouts                              

\usepackage{graphics} 
\usepackage{epsfig} 
\usepackage{mathptmx} 
\usepackage{times} 
\usepackage{amsmath} 
\usepackage{amssymb}  
\usepackage{multirow}
\usepackage{color,soul}
\usepackage{algorithm}
\usepackage{siunitx}
\usepackage{algpseudocode}

\usepackage{hyperref}
\usepackage[comma,numbers]{natbib}

\title{\LARGE \bf
Authoring and Operating Humanoid Behaviors On the Fly using Coactive Design Principles
}

\author{Duncan Calvert$^{1,2}$,
Dexton Anderson$^{1}$,
Tomasz Bialek$^{1}$,
Stephen McCrory$^{1,2}$, \\
Luigi Penco$^{1}$,
Jerry Pratt$^{1,2,3}$
and Robert Griffin$^{1,2}$
\thanks{This work was funded through ONR Grant N00014-19-1-2023, NASA Grant No. 80NSSC20M0197, and ARL Cooperative Agreement W911NF-21-2-0241.}
\thanks{$^{1}$The authors are with the Florida Institute for Human and Machine Cognition, 40 S Alcaniz St, Pensacola, FL 32502, United States}%
\thanks{$^{2}$The author are with the University of West Florida, 11000 University Pkwy, Pensacola, FL 32514, United States}%
\thanks{$^{3}$The author is with Figure AI, Inc., Sunnyvale, CA, United States}%
\thanks{Email : \url{{dcalvert, danderson, tbialek, smccrory, lpenco, jpratt, rgriffin}@ihmc.org}
}} 

\begin{document}

\maketitle
\thispagestyle{empty}
\pagestyle{empty}

\begin{abstract}
Humanoid robots have the potential to  perform useful tasks in a world built for humans.
However, communicating intention and teaming with a humanoid robot is a multi-faceted and complex problem.
In this paper, we tackle the problems associated with quickly and interactively authoring new robot behavior that works on real hardware.
We bring the powerful concepts of Affordance Templates and Coactive Design methodology to this problem to attempt to solve and explain it.
In our approach we use interactive stance and hand pose goals along with other types of actions to author humanoid robot behavior on the fly.
We then describe how our operator interface works to author behaviors on the fly and provide interdependence analysis charts for task approach and door opening.
We present timings from real robot performances for traversing a push door and doing a pick and place task on our Nadia humanoid robot.

\end{abstract}

\section{Introduction}
\label{introduction}

The humanoid form has uniquely diverse mobility and manipulation capabilities that drive its suitability as the embodiment of a general purpose robot.
This has lead to the pursuit of building humanoid robots to perform useful tasks in spaces designed for humans.
There are a number of promising humanoid robot platforms in the world today\cite{bostondynamics2023atlas, promat2023recap}, but humanoid robots that are economical and general purpose are likely still a decade or more away.
One component that is missing is the ability to quickly and effectively get the robot to perform useful tasks with a minimal amount of human supervision, which we refer to as behavior authoring.
This paper explores techniques that build upon known, useful principles in the literature in an effort to nudge the state of the art of behavior authoring on humanoid robots forward.

Behavior \textit{authoring} is the process in which a human operator assembles a system of actionable instructions for the robot to execute a task.
When authoring is concluded, the robot should possess the ability to perform that task in an automatic fashion with high reliability.
In this work, we focus on an interface for authoring behavior ``on the fly" and the interdependence between the operator and robot during the behavior authoring process.
We use the phrase ``on the fly" in this context to mean that the operator is able to create, modify and execute task components while the robot is powered on and in the field.

\begin{figure}[!t]
\centering
    \includegraphics[width=1.0\columnwidth]{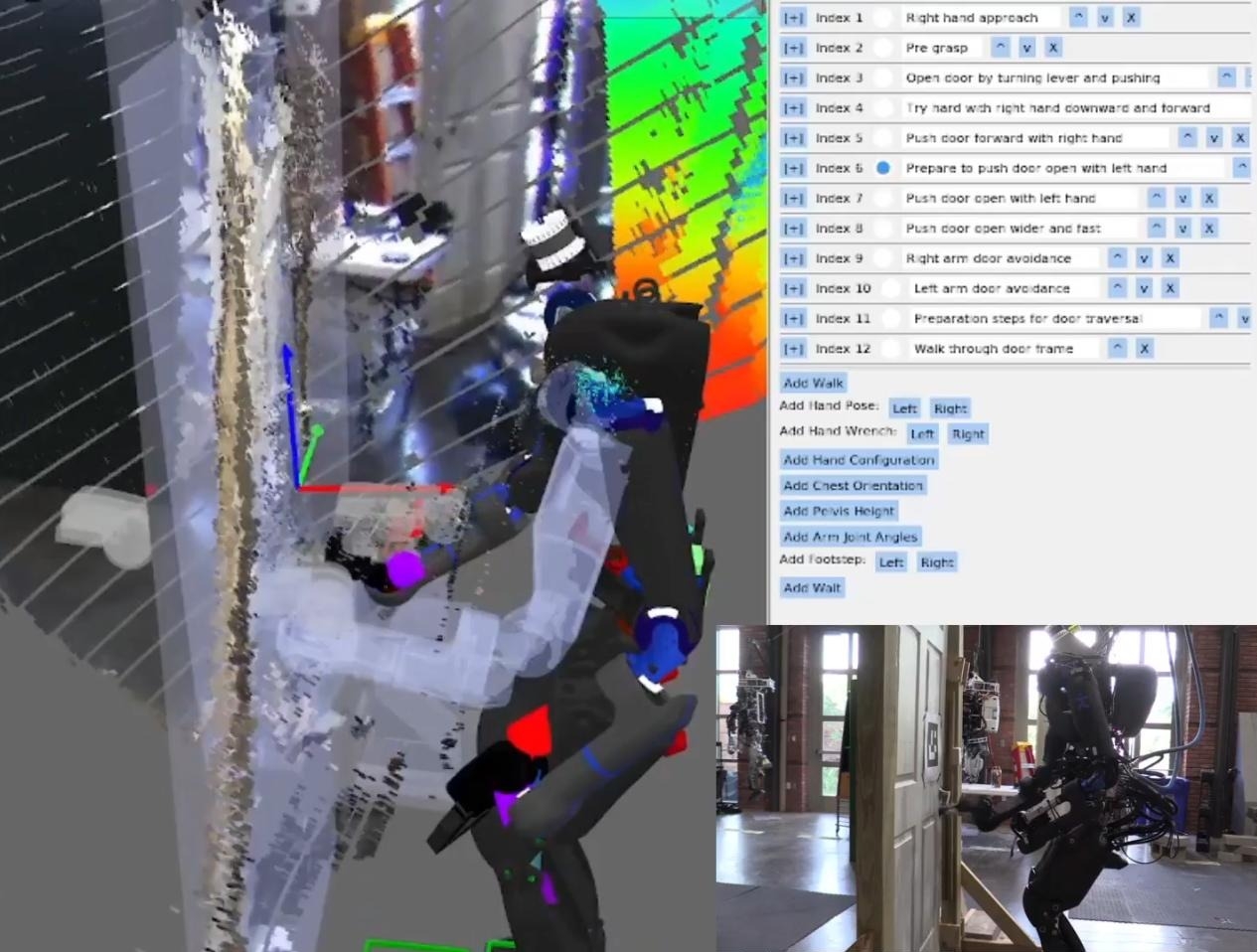}
\caption{Boardwalk Robotics and IHMC's Nadia humanoid robot automatically executing a push door traversal.}
\vspace{-6mm}
\label{fig:push_door_sequence}
\end{figure}

Our system is a new implementation inspired by affordance template architecture in which we have strived to provide a foundation for providing powerful interdependence during operation.
We try to tighten the feedback loop of experimentation and validation in the authoring process, facilitating authoring speed and enabling exploration of a larger set of possible solutions in a realtime setting.
We present interdependence analysis charts for two common actions: task approach and door handle manipulation.
This analysis allows us to formally zoom in on the interactions between the robot and the operator during the authoring process.
We demonstrate the potential of this framework by performing and presenting successful trials on real robot hardware.
In this paper, we present the following contributions:
\begin{enumerate}
    \item A description of key interface elements and how they help the operator interact with the robot.
    \item Discussion of how our operator interface works to author behavior on the fly.
    \item Interdependence analysis charts for task approach and door opening.
    \item Timings of real robot authoring and subsequent automatic execution for pick and place and door traversal.
\end{enumerate}

\section{Humanoid Behavior Authoring}

We abstract humanoid behavior into two primary categories: mobility and manipulation.
\textit{Mobility} is getting somewhere and \textit{manipulation} is doing something once you get there.
We build a hierarchy of abstractions as we dive into each.

Mobility itself has several fields of research that comprise it, including collision-free path planning\cite{Lin_2021}, contact sequence planning\cite{Griffin_2019}, dynamic motion planning\cite{Egle_2022}, and balance control\cite{Koolen_2016}.
Manipulation also has entire fields of research associated with it, which include inverse kinematics\cite{Beeson_2015}, collision-free trajectory planning\cite{MoveIt_2019}, grasp generation\cite{mousavian20196dof}, and semantic planning\cite{driess2023palme}.
Additionally, for autonomy, mobility and manipulation depend on perception, which is a mature research field in its own right. Topics in perception include computer vision, sensor design, semantic segmentation, object pose estimation, SLAM, and scene graphs.

In this work, we focus on navigation through operator placed stance pose goals and assume that sufficient planning and control is available for execution.
Likewise, for manipulation we focus only on inverse kinematics to achieve hand poses and ``open" and ``close" hand configurations.
For perception, we provide the operator with a view of a colored point cloud and reference frames for virtual scene graph objects which are detected using fiducial markers.
We chose these fundamental elements to provide a basic framework and basis for future expansion.

The authoring process should allow the operator to quickly construct new behavior that can later be run in an automatic but optionally supervised mode that executes at near human speed.
The key and qualitative and quantitative measurements of value are:
\begin{enumerate}
  \item The time taken by the operator to author a given task.
  \item The complexity and usefulness of the task.
  \item The degree of human assistance required during execution after authoring.
  \item The robustness and reliability or success rate of the behavior.
\end{enumerate}

In this paper we focus on two basic behaviors of humanoid robots that are fundamental for doing useful work: door traversal and pick and place.

\subsection{Door Traversal}

Door traversals illustrate complexity in mobility.
Traversing doors is not in itself useful, but a means to accomplishing something else.
It is also a task that is uniquely suited to the humanoid form -- most doors are designed for humans.
Doors have handles that are relatively high and designed for human hands to manipulate.
It is also beneficial to use two arms to traverse a door: one to pull or push it open and the other to keep it open.
Door frames are relatively narrow spaces which could require a humanoid to turn sideways where large wheeled or multi-legged bases cannot.
In this paper we focus on a common type of door with a lever handle on one side, a hinge on the opposite side, that swings open only one way, and does not have a automatic closer.
For this type, a push side traversal is generally comprised of the following parts:
\begin{enumerate}
  \item Approach the door near enough to reach the handle.
  \item Grasp and turn or push on the handle enough to disengage the latch.
  \item Push the door open.
  \item Walk through the door.
\end{enumerate}

Traversing the door from the pull side is more complex because you must avoid the door swinging towards you \textit{and} temporarily hold the door open with the hand or arm opposite the hinge side first before transferring that role to the other hand or arm.

\subsection{Pick and Place}

A pick a place task is among the simplest of useful manipulation tasks.
We do not add any further complexity to this task in this work, keeping it to a set of straightforward steps:
\begin{enumerate}
  \item Identify the pose of the object.
  \item Approach near enough to reach the object.
  \item Grasp and lift up the object.
  \item Place the object somewhere else.
\end{enumerate}


\section{Related Work}

The Affordance Template (AT) framework\cite{Hart_2014, Hart_2015, Hart_2022} is a primary inspiration for this work.
It provides an integrated environment for authoring templates for tasks and provides a general definition language for robot-agnostic manipulation.
Examples of ATs include interactive solutions for humanoid robots to pick and place items, operate industrial valves, and open a car door to retrieve an object.
ATs provide support for advanced planning such as stance and grasp generation, navigation and motion planning, and motion primitives to abstract common physical manipulation interactions.
Work on ATs have not explored on the fly authoring.


MIT's Director, developed for the DARPA Robotics Challenge (DRC), used affordance template concepts, a rebuilt framework to integrate robot autonomy components, and a task execution framework\cite{Marion_2017}.
It was used very successfully to score 16 points in the finals.
It includes an operator in the loop pipeline for executing actions, an intuitive 3D scene with interactive widgets, and a way to make custom panels of widgets.
An embedded Python programming environment was used to write task scripts, however, the authoring process is not detailed.

mc\_rtc is an integrated framework for managing robot behavior that supports bringing your own robot model as a URDF.
It allows the user to write behaviors that can be run on real robot and simulation using the same interface\cite{Singh_2023}.
It has been used to get humanoid robots walking up stairs, driving vehicles, and performing industrial manipulation tasks.
The framework allows the developer to programmatically construct finite state machines that can then be operated by the user interface.
However, mc\_rtc does not have an interactive behavior authoring interface.

\begin{figure}[h]
\centering
    \includegraphics[width=0.7\columnwidth]{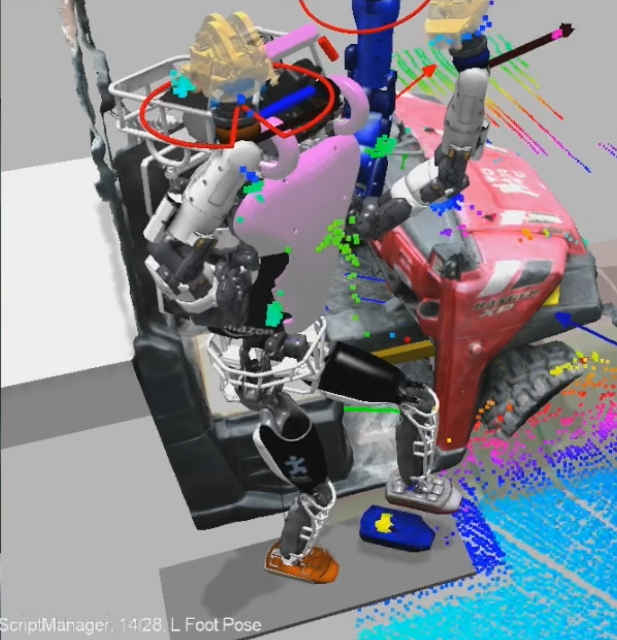}
\caption{In the 2015 DRC UI, when an action is selected for execution, a preview of the action's goal state is shown situated in the 3D view. An interactable footstep is shown in blue, representing the next action. This virtual object can be clicked on by the mouse and adjusted using gizmos before being executed by the operator, providing observability, predictability and directability.}
\vspace{-1mm}
\label{fig:drc_egress}
\end{figure}

The Coactive Design method\cite{Johnson_2014} is an iterative process comprised of three main processes: an identification process, a selection and implementation process, and an evaluation of change process.
The most complex process is in the identification process in which requirements, alternatives, and interdependence relationships are explored.
A set of desired interdependence relationships are determined and selected for implementation.
The result is then evaluated using human feedback and performance analysis.
This method was used for the design and development of the operator interface used by IHMC in the 2015 DRC\cite{Johnson_2014}.
A later analysis details how that methodology led to success in the competition\cite{Johnson_2017}.
The car egress shown in \autoref{fig:drc_egress} used a scripting engine which allowed the operator to cycle through a sequence of predefined actions.
This scripting engine is a precursor to the presented work in this paper, which has been rewritten with heavy reference to the original and applies lessons learned at a base architectural level.


\section{Authoring Interface}


Our interface is designed to put the operator and robot in a situation where they are engaged in rich interaction with data.
It is an environment in which behaviors can be created from scratch, existing behaviors modified, and end-to-end tested while the robot is powered on and performing action in the field.
We refer to this as ``on the fly" authoring because it can be used to accomplish and automate tasks as they are encountered.
Achieving this requires elements to have observability, predictablilty, and directability.

\begin{figure}
\centering
    \includegraphics[width=1.0\columnwidth]{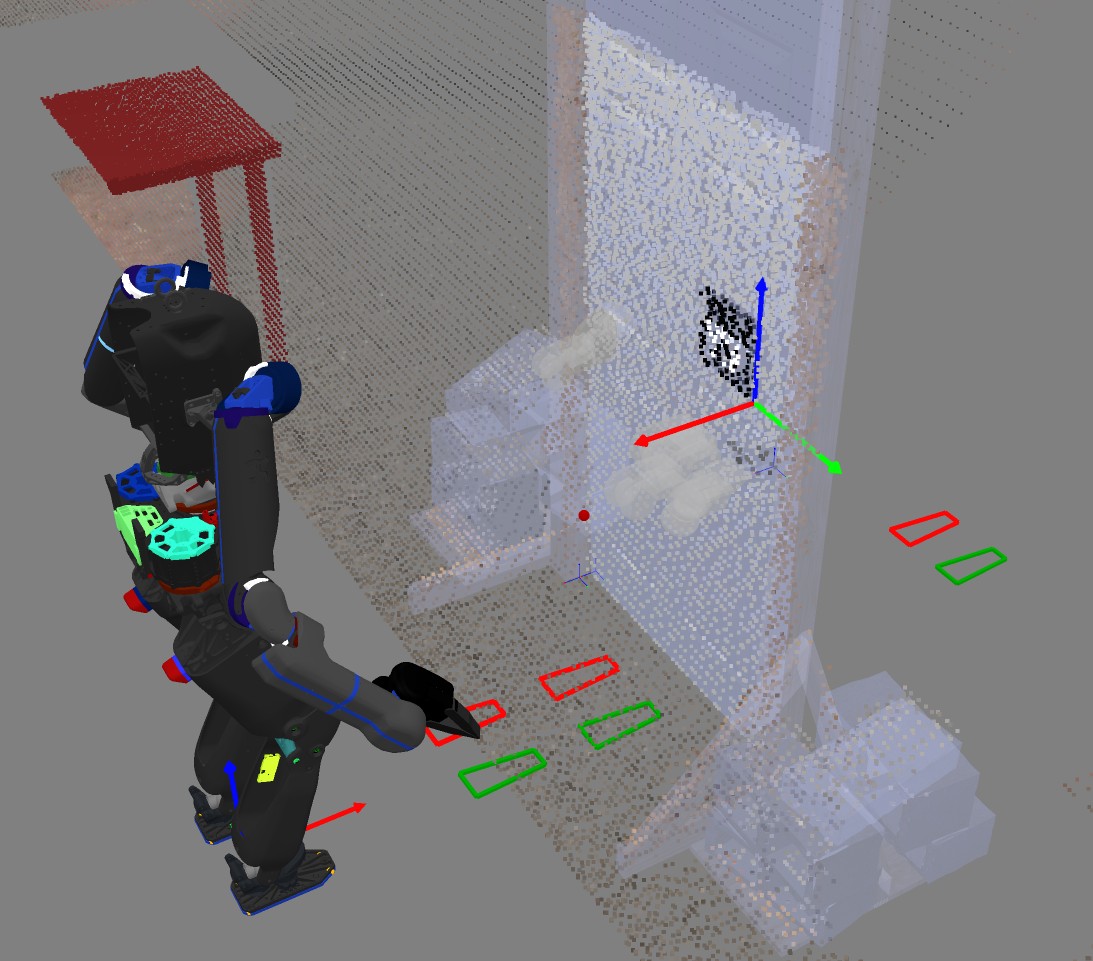}
\caption{The push door behavior is selected, shown situated in the 3D scene, as the robot is facing the door.}
\vspace{-2mm}
\label{fig:push_door_behavior_layout}
\end{figure}

A 3D scene is the central focus, where behavior keyframes and widgets are laid out in world space, as seen in \autoref{fig:push_door_behavior_layout}.
The user can orbit the camera and move the focus point with the mouse and keyboard.
Visualizing behavior data in 3D allows the operator to inspect spatial relationships which is a fundamental part of verification.
If this information was not visualized, the operator would be forced to doubt the validity and intention of the behavior.
They would need to hold in mind questions like ``Is the next hand pose where I think it is?" and ``Where does the robot end up at the end of the task?".
Sometimes, there are bugs and issues where the task actions would be completely somewhere else in the world.
With this approach, it is intuitive for the operator to verify the alignment of task actions to the task.


\begin{figure}[h]
\centering
    \includegraphics[width=0.6\columnwidth]{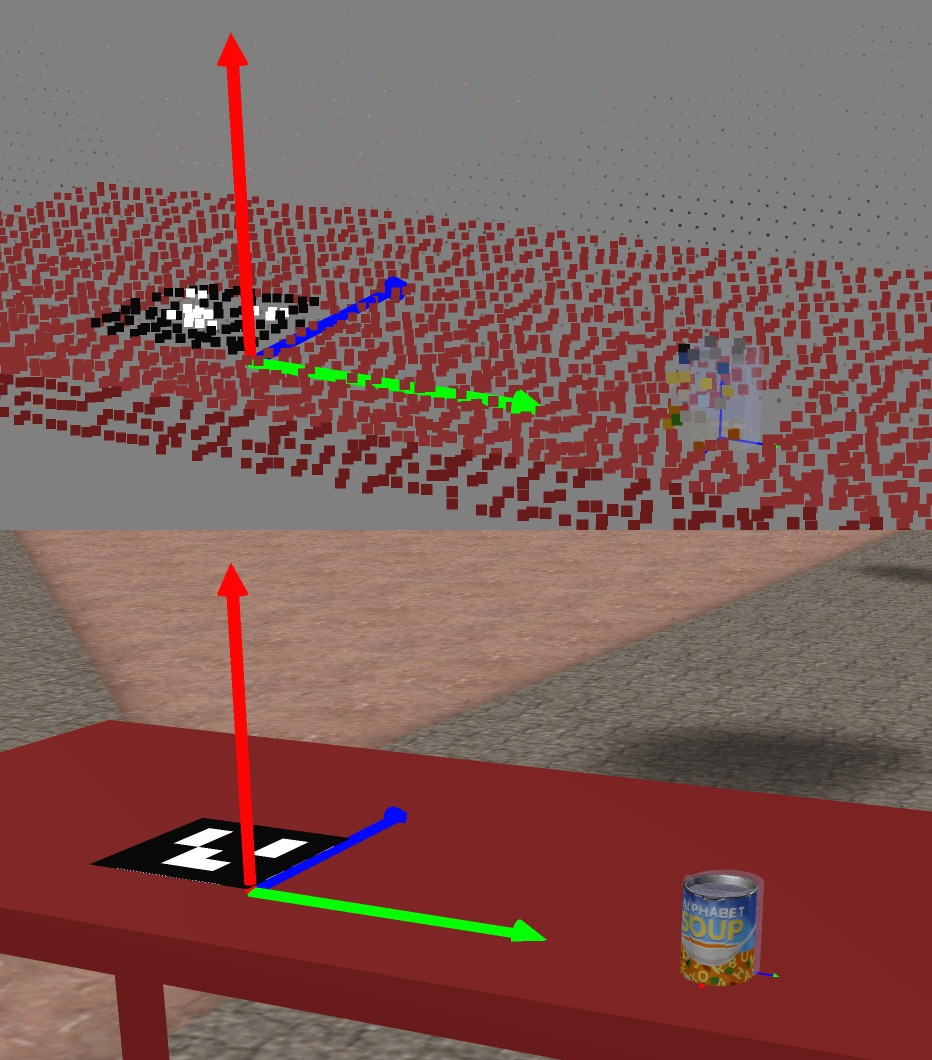}
\caption{A can of soup detected by an ArUco marker in simulation.
The large coordinate frame shows the detection of the ArUco marker and the small blue coordinate frame shows the can of soup task frame which behavior action can be authored with respect to.
The top image shows a point cloud representation that the operator would see and the bottom shows the ground truth for clarity.}
\vspace{-2mm}
\label{fig:can_of_soup_detection}
\end{figure}

Interacting with parts of a behavior in a 3D setting provides important context when overlayed with a model of the environment.
For example, authoring a hand pose in code would have a name, description, and numbers associated with it, but seeing the pose and hand mesh graphic in 3D shows the proximity to other things in the environment.
The hand pose may be colliding with the surface of a table, or it might be very near a wall, which may be relevant to the planning complexity of the surrounding action keyframes.
The operator will notice these 3D spatial relationships and constraints and may choose to modify the behavior based on it.

Our model of the environment consists of colored point clouds and predefined detected objects such as doors and cups as shown in \autoref{fig:can_of_soup_detection}.
These get rendered in the 3D scene for the operator to view from different perspectives.
Detected objects have reference frames which can be used to specify action poses with respect to that object.
Tasks are defined with respect to predefined task frames such that the behavior can occur anywhere in the world that task is encountered.

\begin{figure}
\centering
    \includegraphics[width=0.8\columnwidth]{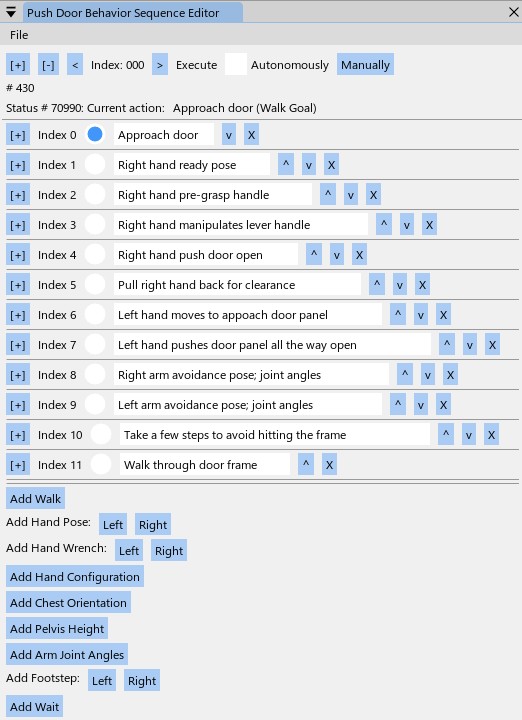}
\caption{The editor panel that facilitates authoring a scripted linear sequence of behavior.
The checkbox at the top is used to direct the robot to proceed with the actions automatically.
Next to it is a button named "Manually" that is used to execute only the currently selected action.}
\vspace{-2mm}
\label{fig:sequence_editor}
\end{figure}

To keep the operator in the loop and able to respond to more scenarios, direct teleoperation tools are kept readily available in the same application.
The tools include include manual footstep placement and upper body kinematic streaming using virtual reality controllers.
They can be used to take over task performance, recover from failures, or avoid damage to the robot by getting it out of tricky situations.


To create, edit, and execute actions, an action sequence editor is used, shown in \autoref{fig:sequence_editor}.
It represents a linear sequence of execution.
Actions can be added through the buttons at the bottom and tuned using cooresponding panels of widgets specific to that action.
The action panels are lined up in order from top to bottom.
The action panels can be expanded and minimized, shown in the figure minimized.
Each action is given a hand written human-readable description, which is important for remembering the semantic context of that action.

\begin{figure}
\centering
    \includegraphics[width=1.0\columnwidth]{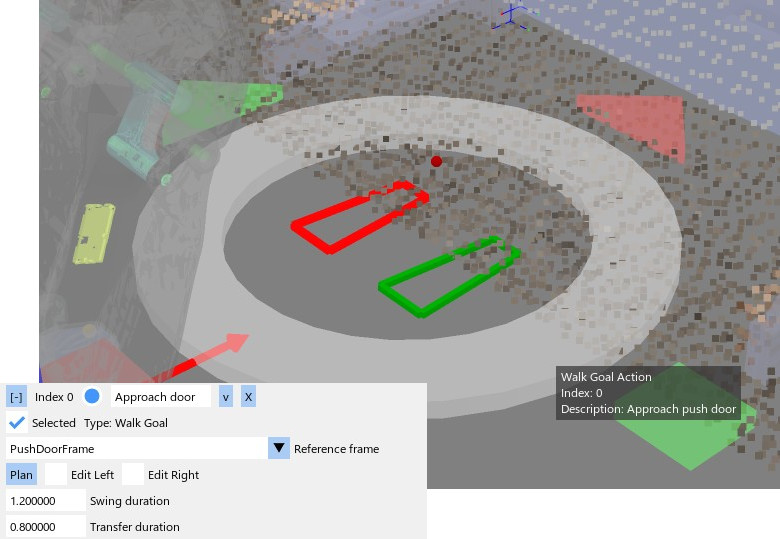}
\caption{A stance pose action situated in the 3D view. 
It represents the goal stance with reference to a task frame for a footstep planner to plan to.
The panel that is used to configure a stance pose is also shown.}
\vspace{-2mm}
\label{fig:stance_pose_action}
\end{figure}

\begin{figure}
\centering
    \includegraphics[width=1.0\columnwidth]{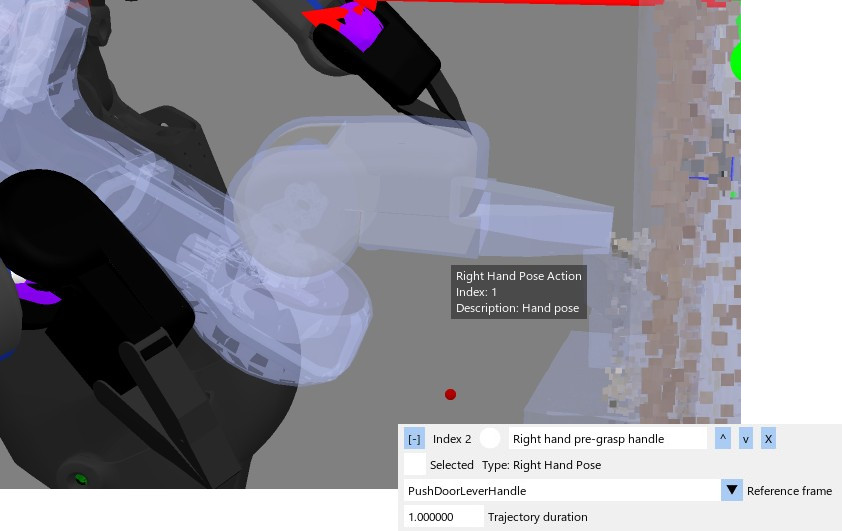}
\caption{A hand pose action situated in the 3D view. 
It represents a goal pose for the inverse kinematics solver to solve for joint angles to achieve.
A ghost preview of the arm's kinematic solution is previewed as it is moved around.
A panel for adjusting the properties for a hand pose action is also shown which allows operator can adjust the parent frame and the trajectory duration.}
\vspace{-2mm}
\label{fig:hand_pose_action}
\end{figure}

\begin{figure}
\centering
    \includegraphics[width=1.0\columnwidth]{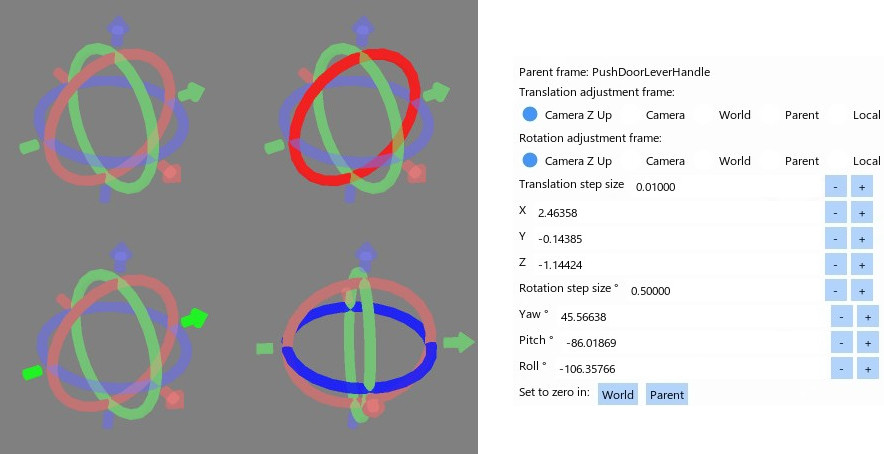}
\caption{A gizmo for manipulating a 6 DoF pose with a mouse and keyboard.
The mouse can be used to drag the rings and arrows to perform constrained adjustments to those axes.
The keyboard arrow keys can be used to move the gizmo with respect to the adjustment frames selected in the context menu.
Modifier keys Ctrl, Alt, and Shift are used to select rotation axes and adjust the sensitivity.
The context menu allows the user to change the frame of adjustment, nudge the pose by small amounts, and view and edit the abolute values of the pose.}
\vspace{-2mm}
\label{fig:pose_gizmo}
\end{figure}


\begin{figure*}[tbh!]
\centering
    \includegraphics[width=1.7\columnwidth]{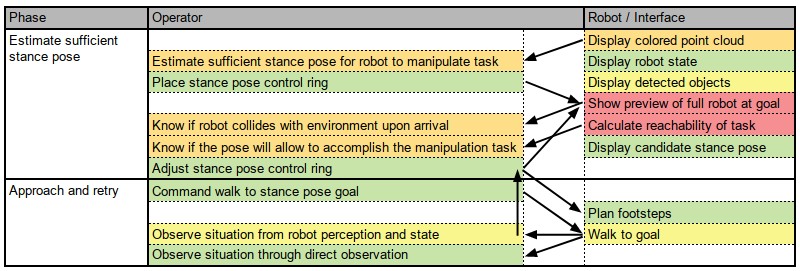}
\caption{Interdependence analysis of a basic manipulation task approach.}
\vspace{0mm}
\label{fig:task_approach_ia}
\end{figure*}

To specify a stance pose goal, we use a ring graphic with footstep outlines in the middle as shown in \autoref{fig:stance_pose_action}.
It is used to specify where the robot should be standing at some phase in the task.
It can be translated on it's X-Y plane by dragging the ring with the left mouse button and yawed by dragging on the ring with the right mouse button.
The red and green arrows are used to specify which way the X forward and Y left axes are facing and to quickly reorient the pose.
Clicking the arrows will orient the ring to face that direction.
A tooltip gives more information about the stance pose when the mouse hovers it.
A panel is also provided which can be used to adjust parent reference frame, swing and transfer duration of the planned steps\cite{Griffin_2019}, and the poses of the stance steps individually in order to achieve a staggered pose.

\begin{figure}
\centering
    \includegraphics[width=1.0\columnwidth]{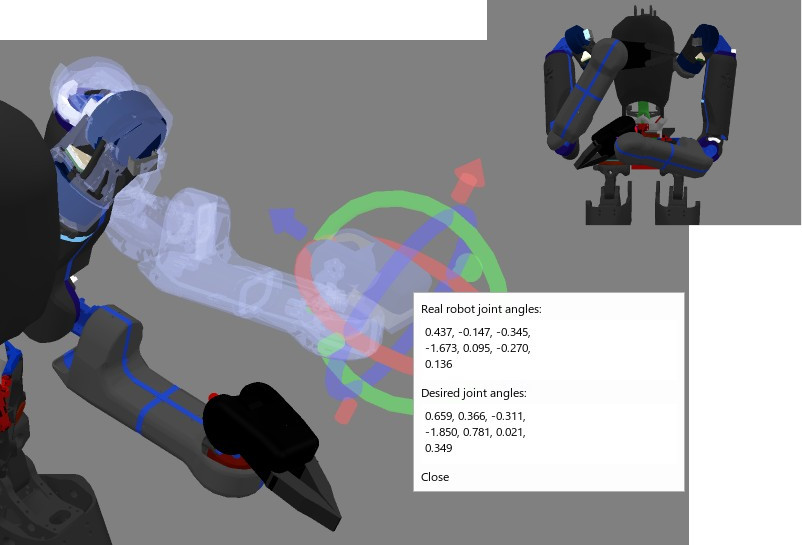}
\caption{The configuration on Nadia's arms at the collision avoidance joint angles.
We implement a tunable action that will move the arms to a set of joint angles.
Right clicking the interactable hand in the 3D scene shows the robot's current joint angles and the joint angles solution of the inverse kinematics solver from moving the virtual hand with the gizmo.}
\vspace{-2mm}
\label{fig:collision_avoidance_arms}
\end{figure}

To specify a hand pose goal, we use a semi-transparent model of the hand as shown in \autoref{fig:hand_pose_action}.
It is used to specify where the hand should be in at some point during the behavior.
When selected, the pose gizmo shown in \autoref{fig:pose_gizmo} appears around it and can be used to control X, Y, Z translation and yaw, pitch, and roll rotation.
When fine or absolute adjustment is needed, a context menu can be used by right clicking the gizmo.


When the robot needs to walk through a tight space like a door frame, we want to pull the arms into a specific configuration for that.
The arms configuration for that on Nadia is shown in \autoref{fig:collision_avoidance_arms}.
Joint angle based arm configurations can be more reliably executed because they don't rely on the inverse kinematics solver to give a consistent solution for all joints along the arm.
This is important for pulling the arms in because the configuration is carefully tuned for the arms to take up as little space as possible.
The context menu available by right clicking on an interactive hand in the 3D scene is used to gather specific values.





\section{Interdependence Analysis}

In this section we zoom in on the interdependence relationships between the operator and the robot for two key sub-tasks: approaching a task and opening a door.
The purpose of interdependence analysis is to identify key parts of a process where the human and the robot are interacting and depending on each other.
It also used as a tool to find key shortcomings in the system by analyzing the reliability and capability of task components with relation to their dependencies.
The leaf nodes are color coded to show where things can go wrong and where they usually go right.
Using green, yellow, orange, and red we show the gradient from ``sufficient" to ``not functional".
Arrows are drawn to indicate dependencies and the flow of information.
Where a dependency is yellow, orange, or red are where improvements to the system's reliability and capability can be made.
Green indicates that something generally works well.
Yellow indicates that the reliability or capability is less than 100\%.
Orange indicates that the reliability or capability is not sufficient, causing dependencies to suffer.
Red indicates missing functionality or something that never works.


\subsection{A Task Approach Scenario}

\begin{figure*}[htb!]
\centering
    \includegraphics[width=2.0\columnwidth]{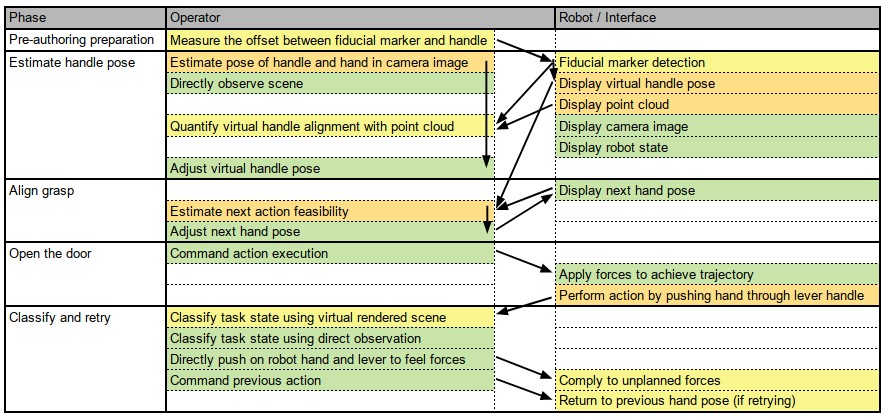}
\caption{Interdependence of opening a door by disengaging the latch via a lever handle.}
\vspace{0mm}
\label{fig:door_latch_ia}
\end{figure*}

Our interdependence analysis chart for authoring the approach of a manipulation task can be seen in \autoref{fig:task_approach_ia}.
It is consists of a table with columns for phase, operator, and robot and colored cells that represent the components.
The first phase is observing the situation and estimating a good approach location for the robot.
The operator uses the colored point cloud, robot state visualization (which is just a graphic of the robot at some pose and configuration), and ghost objects that are detected by the perception sensors.
However, the colored point cloud is not very detailed.
Our point cloud does not currently use filters over time or construct an accurate mesh of the environment -- it shows a relatively noisy scattering of points.
Because of this, its difficult to tell where things are, and introduces uncertainty for the operator in estimating a good approach.
When approaching a task, you don't want the robot to bump into a table or wall or disturb items that you may want to grasp.
We show in red that there are two currently missing functionalities that would improve on this.
If we rendered a previewed robot at the goal position, the operator could check visually for collisions.
Another goal of task approaches in general is to arrive at a stance such that the robot can reach what it needs to without having to readjust.
We do not currently have a reachability analysis tool to help with that, so we also show it in red.

When the operator is happy with the candidate stance pose they adjusted with the control ring, they command the robot to walk there.
Planning footsteps is usually not an issue here so we show it as green.
Because we are uncertain if the arriving at the stance pose will result in a collision, the robot may bump into the task and fall.
Once the walking has completed, the operator looks at the 3D scene view to qualify the result.
They may also choose to look at the robot and environment directly (in the case of local operation).
The operator may choose to further adjust the stance goal, command it, and repeat until the result is sufficient.

\subsection{Opening a Door}

\autoref{fig:door_latch_ia} shows an interdependence analysis chart for opening a door with a lever handle.
This task is aligned with a lever handle which is preregistered as a scene object with a static transform relative to an ArUco marker.
We show that this is somewhat problematic as we have had difficulty getting a good measurement of that transform.
The operator can see a 2D camera video feed from the robot's head but that is not sufficient to estimate a pose with accuracy.
In this case, the operator was forced to directly observe the scene because of these shortcomings in dependencies.
Then, the operator adjusts the grasp hand pose while estimating the feasibility that the motion will successfully disengage the latch.
The operator cannot know this will work, so we color it orange.
The operator commands the action and the robot applies forces on the world and may push through the lever and forward enough to open the door.
After this, the operator must classify the new state as successful or failed.
In the case of failure, we illustrate that the operator may alternatively approach the robot and push on the lever and robot's hand a bit to get a sense for the forces involved and adjust the hand motion accordingly.
The operator will iterate on this process until success is achieved.


\section{Results}

In \autoref{tab:pick_and_place_authoring}, we show the key timestamps in authoring a pick and place action sequence on the real robot.
Each timestamp after the first represents the time in which that action was completed.
In this run the operator took 38 minutes to author actions on the fly such that the robot had picked up the can, stepped to the side, and placed it down again during the authoring process itself.
\autoref{tab:pick_and_place_execution} shows the timestamps for re-executing the authored behavior step-by-step while supervising the robot's progress.
The execution was around 14 times faster than the authoring.
\autoref{tab:push_door_traversal_authoring} shows the timestamps of the authoring process of a push door traversal which took 33 minutes in which the robot performed the task succesfully during authoring.
\autoref{tab:push_door_traversal_automatic} shows the timestamps of executing the push door traversal in fully automatic mode in which the operator did not intervene after the initial command.

\begin{table}
\caption{Authoring pick and place of a can of soup.}
\centering
\begin{tabular}{l l}
 \hline
 Time (m) & Action authored \\
 \hline
 0 & Create new action sequence \\
 2 & Rough manual table approach \\
 3 & Approach table \\
 4 & Right hand approaches can \\
 11 & Pre-grasp pose sufficient \\
 12 & Robot grasps can of soup \\
 13 & Robot lifts can of soup \\
 22 & Pull arm back action setup \\
 23 & Side step action executed \\
 28 & Put arm forward \\
 32 & Set down can of soup \\
 36 & Release grasp \\
 37 & Pull arm back \\
 38 & Back away from table \\
\label{tab:pick_and_place_authoring}
\end{tabular}
\end{table}

\begin{table}
\caption{Step-by-step supervised execution of picking and placing can of soup.}
\centering
\begin{tabular}{l l}
 \hline
 Time (m:s) & Action completed \\
 \hline
 0:00 & Begin approach \\
 0:11 & Approach table \\
 0:14 & Right hand approaches can \\
 0:53 & Pre-grasp hand pose \\
 0:58 & Grasp can of soup \\
 1:00 & Pull back hand with can of soup \\
 1:16 & Step to the side \\
 1:20 & Set down can \\
 1:36 & Release grasp on can \\
 1:46 & Back away from task \\
\label{tab:pick_and_place_execution}
\end{tabular}
\end{table}

\begin{table}
\caption{Authoring push door traversal.}
\centering
\begin{tabular}{l l}
 \hline
 Time (m) & Action authored \\
 \hline
  0 & Create new action sequence \\
  2 & Approach door \\
  5 & Right hand approaches handle \\
  7 & Pre-grasp hand pose \\
  9 & First handle turn contact \\
 14 & Latch disengaged \\
 20 & Door pushed open with right hand \\
 24 & Door pushed open more with left hand \\
 25 & Door pushed open all the way with left hand \\
 26 & Arms in collision avoidance configuration \\
 32 & Step forward a little \\
 33 & Walk through the door frame \\
\label{tab:push_door_traversal_authoring}
\vspace{-3mm}
\end{tabular}
\end{table}

\begin{table}
\caption{Fully automatic push door traversal.}
\centering
\begin{tabular}{l l}
 \hline
 Time (m:s) & Action completed \\
 \hline
 0:00 & Begin approach \\
 0:07 & Approach door \\
 0:11 & Right hande approaches handle \\
 0:12 & Pre-grasp \\
 0:14 & Door opened \\
 0:20 & Push door all the way open with left hand \\
 0:24 & Arms in collision avoidance configuration \\
 0:35 & Finish walking through door frame \\
\label{tab:push_door_traversal_automatic}
\vspace{-6mm}
\end{tabular}
\end{table}

\begin{figure}
\centering
    \includegraphics[width=1.0\columnwidth]{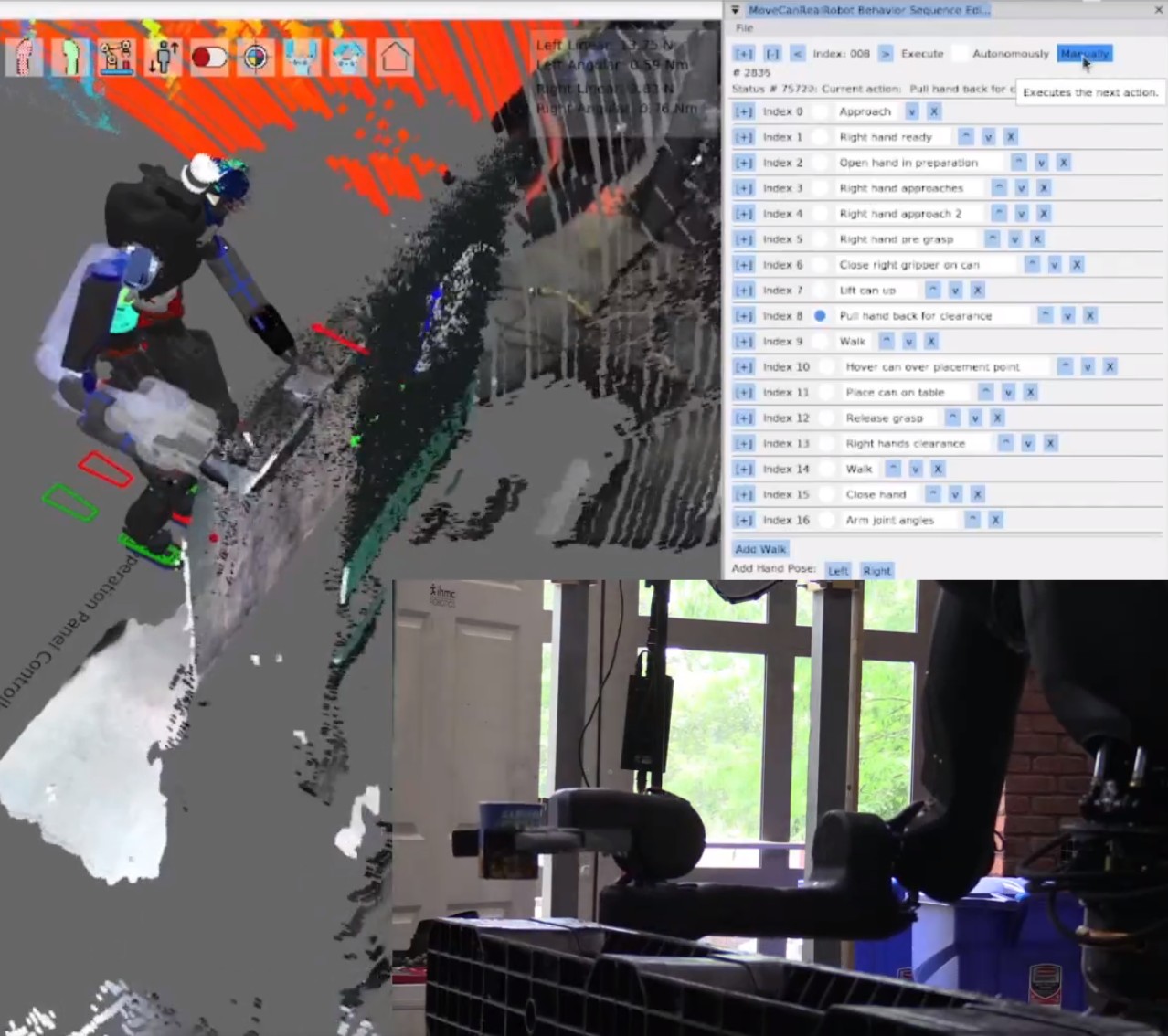}
\caption{The Nadia humanoid robot executing the pick and place action sequence for a can of soup.}
\vspace{0mm}
\label{fig:can_pick_and_place}
\end{figure}

\section{Discussion and Future Work}

We think that this work provides a base for expansion.
There are a lot of known improvements to the user interface details that would make the authoring process faster.
One of the most significant developmental difficulties is that many issues only appear when running the full setup on the real robot, thus making it difficult to document and address issues.


There are algorithms for stance and grasp generation available as detailed in \cite{Hart_2022} which could be used to automate task approach, considering reachabilty, and to plan grasps on modelled objects.
Today's neural net assisted algorithms\cite{tremblay2018deep}\cite{Lin_2022} could enable the detection of a wide variety of objects and remove the need for fiducial markers.
Having an expansive library of detectable objects would open up the space for autonomous behavior.
There is a growing area of research termed ``affordance primitives"\cite{Pettinger_2020}\cite{Pettinger_2022} which are a way to model actions with constrained movement, such as turning a valve or closing a drawer.
Affordance primitives have the potential for force and perceptual feedback control during the manipulation task.
In general, the authoring process should evolve into higher level abstractions, such as specifying the desired poses of objects rather than the movement of the robot to get them there.


\section{Conclusion}

In this work, we developed a new interface for authoring humanoid behaviors with the ability to do the authoring on the fly.
We detailed several interactive interface elements and how they are used in the authoring process.
We introduced interdependence analysis charts for a few key tasks that humanoid robots encounter.
Finally, we demonstrated this approach by conducting two experiments on real robot hardware and provided the timings of the authoring process and execution processes.

\subsection{Acknowledgements}

We would like to thank Stephen Hart, Matt Johnson, William Howell and the team at IHMC Robotics, without whom this work would not have been possible.

\subsection{Source Code and Media}

Our implementation X and the associated modules discussed in this paper can be found on our GitHub at \url{https://github.com/ihmcrobotics}.
The accompanying video can be found at \url{https://youtu.be/SbBGpRHY_eE}.

\bibliography{mybib}

\end{document}